\documentclass[final,11pt,times]{elsarticle}

\usepackage{amssymb}
\usepackage{geometry}
\usepackage{graphicx}
\usepackage{booktabs}
\usepackage{tabularx}
\usepackage{array}
\usepackage{fancyhdr}
\usepackage[colorlinks=true, pdfauthor={Zhichao Zhou}]{hyperref}

\newcommand{\maybeincludegraphics}[2][]{%
  \IfFileExists{#2}{\includegraphics[#1]{#2}}{\fbox{\scriptsize\ttfamily\detokenize{#2}}}%
}

\newcolumntype{Y}{>{\raggedright\arraybackslash}X}

\journal{CIE53 Proceedings}

\begin{document}

\begin{frontmatter}

\title{Retrieval-grounded robot program generation and simulation-based correction via Model Context Protocol}

\author[labela]{Zhichao Zhou}
\ead{zhousaiensi@gmail.com}
\author[labela]{Siyuan Chen\corref{cor}}
\ead{siyuan.chen@chalmers.se}
\cortext[cor]{Corresponding author}
\author[labela]{Omkar Salunkhe}
\ead{omkar.salunkhe@chalmers.se}
\author[labela]{Ebru Turanoglu Bekar}
\ead{ebrut@chalmers.se}
\author[labela]{Johan Stahre}
\ead{johan.stahre@chalmers.se}
\author[labela]{Anders Skoogh}
\ead{anders.skoogh@chalmers.se}
\address[labela]{Department of Mechanical Engineering, Chalmers University of Technology, Hörsalsvägen 7a, Gothenburg, SE-412 96, Sweden}

\begin{abstract}
Flexible manufacturing requires industrial robots to be reprogrammed rapidly as product variants change. This paper presents a language-model-based workflow that generates, validates, and iteratively corrects ABB RAPID robot programs from natural language task descriptions. A dual-stream retrieval-augmented generation (RAG) pipeline grounds code generation in verified technical documentation and production templates, reducing domain-specific errors produced by ungrounded language models. A custom Model Context Protocol (MCP) server connects the language-model client directly to ABB RobotStudio for automated code upload, simulation execution, and diagnostic feedback. The evaluation combines a 30-query retrieval benchmark, scoped code-generation checks, and RobotStudio case studies in a simulated pick-and-place manufacturing cell. The simulation loop exposes execution failures that static and semantic checks alone cannot catch, including suction release-height errors, unreachable placement targets, and configuration-dependent recovery motions. The results show how RAG and MCP can connect grounded code generation with executable feedback from industrial robot simulation software, while reducing but not eliminating expert setup and final supervision.
\end{abstract}


\end{frontmatter}
\thispagestyle{fancy}

\section{Introduction}
\label{sec:introduction}

Flexible manufacturing depends on the ability to reconfigure production systems when product variants, fixtures, tools, or production volumes change \cite{johansson2009enabling}. Industrial robots are central to this flexibility because they can perform handling, assembly, and process tasks across different product families. Yet changing a robot task is still not as simple as changing a product plan: the new behavior must be programmed, checked against the cell layout, and validated before it can be trusted in production \cite{hagele2016industrial}.

Robot programming therefore remains a persistent engineering bottleneck. Conventional online and offline programming methods reduce deployment risk, but they still require specialist knowledge of robot languages, coordinate frames, tools, sensors, and process constraints \cite{pan2012programming}. This challenge becomes more visible in high-mix manufacturing, where product variants change frequently and the cost of repeated manual programming can limit the practical flexibility of the cell.

Large language models (LLMs) offer a new interface for robot programming because they can translate natural-language intent into code-like control logic \cite{liang2023code}. Studies on conversational robot programming show that these models can reason about tasks and robot actions, but they also reveal limitations in reliability and domain specificity \cite{vemprala2024chatgpt}. Retrieval-augmented generation (RAG) addresses part of this problem by grounding generation in external documentation and examples \cite{lewis2020rag}. For robot code generation, this is useful because local manuals and validated examples reduce hallucinated instructions and invalid programming patterns \cite{salunkhe2026rags}.

However, grounded text generation is not enough for industrial robot programming. A generated robot program may follow the requested task and still fail when executed in a configured cell. Reachability, path geometry, gripper release behavior, I/O state, and controller recovery are execution properties, not only code properties. Automated testing research for industrial robotic systems makes the same point from a validation perspective: acceptance must be based on system behavior, not only on program structure \cite{dossantos2024aat4irs}. This leaves a gap between RAG-based code generation and the execution-level feedback needed for manufacturing use.

This paper addresses that gap in a concrete industrial robot programming environment. The workflow targets ABB RAPID programs and uses RobotStudio as the simulation and controller-execution environment. RAG retrieves relevant technical documentation and production templates, while the Model Context Protocol (MCP) exposes RobotStudio operations as callable tools. Generated programs are uploaded to a virtual controller, executed, monitored, and corrected using execution status, event logs, joint readings, I/O signals, scene information, and program variables. Screenshots are retained for documentation and human confirmation, but not used as the primary automated detection signal.

The paper contributes a RobotStudio-connected workflow for generating and validating vendor-specific robot programs from natural language, an evaluation of dual-stream RAG for reducing domain-specific generation errors, and case evidence showing how execution feedback can reveal failures missed by text-level checks. The paper is structured as follows. Section~\ref{sec:literature} reviews related work on industrial robot programming, offline simulation, RAG, and tool-using LLM workflows. Section~\ref{sec:workflow} describes the proposed workflow. Section~\ref{sec:setup} explains the evaluation design and metrics. Section~\ref{sec:results} reports the retrieval, generation, and RobotStudio results, followed by discussion and conclusion.

\section{Literature Review}
\label{sec:literature}

Industrial robot programming is shaped by proprietary controller languages and closed platform ecosystems \cite{pan2012programming}. ABB systems use RAPID modules to define motion instructions, workobjects, tool data, I/O behavior, and execution routines \cite{abbrapid2024}. Offline programming and simulation environments such as ABB RobotStudio allow robot programs to be developed and tested before deployment \cite{abbrobotstudio2024}. CAD-based offline programming has also been studied as a way to reduce manual programming effort and make robot programming more accessible to users with manufacturing process knowledge \cite{neto2013offline}. These approaches are useful because virtual controllers and simulation environments can reproduce controller behavior, robot kinematics, and many execution constraints without immediate access to hardware.

For generated robot programs, offline simulation is more than visualization. Code that passes a structural check can still fail because a motion is unreachable, a joint solution approaches a singularity, an arc is geometrically invalid, or a part is not released by an end-effector. Automated acceptance testing for industrial robotic systems has therefore been identified as an important direction for reducing manual validation effort \cite{dossantos2024aat4irs}. In the context of LLM-generated robot code, this means that text-level checks should be complemented by execution feedback from the target programming and simulation environment.

Retrieval-augmented generation addresses a separate but related problem: LLMs are prone to hallucinating technical details when asked to generate code in specialized domains \cite{zhang2025hallucinations}. RAG injects external knowledge at inference time by retrieving relevant source material and conditioning generation on it \cite{lewis2020rag}. Retrieval quality is especially important in technical domains because relevant source material must appear early in the prompt context to influence generation. Query reformulation methods based on hypothetical documents illustrate one way to reduce the vocabulary gap between user requests and technical documents \cite{gao2023hyde}. For robot code generation, this is useful because users describe intentions such as ``pick a part from the conveyor'', while the knowledge base describes instructions, data types, and controller semantics.

Tool-using LLM workflows extend RAG by adding query decomposition, typed tool calls, and iterative correction \cite{yao2023react}. Recent work on agentic retrieval-augmented generation also combines retrieval with tool use and feedback \cite{singh2025agentic}. The Model Context Protocol provides a standardized client-server pattern for exposing external tools and data sources to LLM applications \cite{anthropic2024mcp}. In the proposed system, MCP is used as the interface layer between the LLM client and RobotStudio. This separation keeps the language model outside the simulator while making simulator actions available as typed tools.

These strands show that AI assistance can support robot-program synthesis, but they leave an important validation gap for industrial cells. Language-model robot control has shown that natural-language intent can be translated into executable-looking programs \cite{liang2023code}. Conversational robot-programming studies report similar potential, while also emphasizing reliability limits \cite{vemprala2024chatgpt}. RAG reduces some domain-specific hallucinations by grounding generation in manuals and examples \cite{salunkhe2026rags}. However, text-level correctness does not establish that a generated RAPID module will behave correctly in a configured RobotStudio station. The novelty of the present work is therefore not only using an LLM to produce robot code, but connecting generation to execution-level feedback from RobotStudio, where controller state, event logs, I/O signals, RAPID variables, and scene objects can be used to diagnose and correct failures.

\section{Proposed Workflow}
\label{sec:workflow}

\begin{figure}[htbp]
\centering
\includegraphics[width=0.8\linewidth]{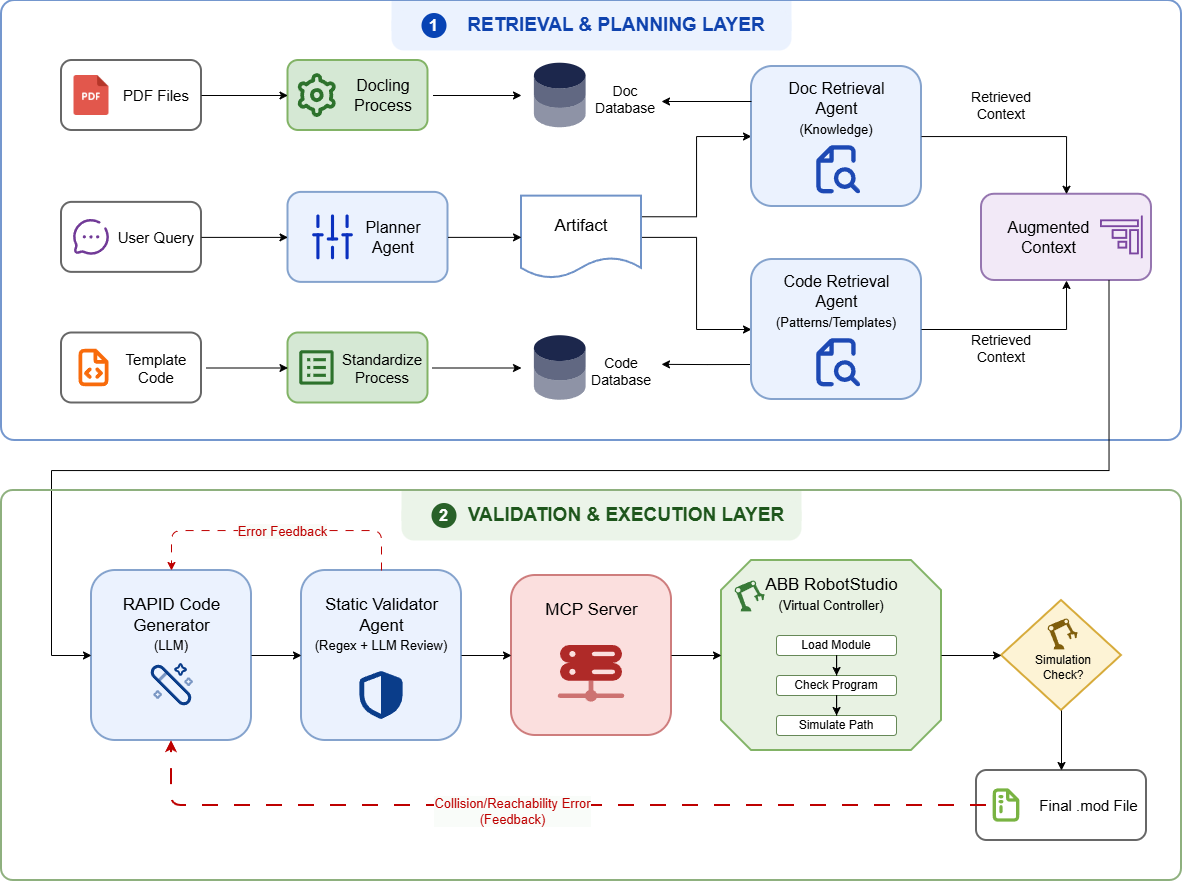}
\caption{Overall retrieval-and-planning and validation-and-execution workflow.}
\label{fig:architecture}
\end{figure}

The workflow has two connected layers: a retrieval and planner layer that prepares grounded RAPID-generation prompts, and a validation and execution layer that runs generated programs in RobotStudio and returns diagnostic feedback. Figure~\ref{fig:architecture} summarizes the overall architecture.

\subsection{Retrieval and planner layer}
\label{subsec:generation}

The retrieval and planner layer uses two retrieval streams. The documentation stream is mainly based on ABB's RAPID technical reference manual for instructions, functions, and data types, while the code-template stream contains 106 validated RAPID examples provided by a Nordic leading vehicle company. This separation is important because the two streams answer different questions. Documentation explains instruction semantics and data types, while templates show how instructions are combined into executable module structures.

For a user request, a query-decomposition step splits the task into smaller retrieval tasks, such as locating motion instruction rules, I/O signal patterns, tool definitions, workobject conventions, or error handling examples. Retrieved snippets are assembled into a structured prompt that requires a complete RAPID module, explicit entry procedure, balanced delimiters, no unresolved placeholders, and use of user-specified tool and workobject names.

\subsection{Validation and execution layer}
\label{subsec:mcp}

The validation and execution layer connects the LLM client to RobotStudio through a RobotStudio MCP bridge developed for this study and released as an engineering artifact.\footnote{The RobotStudio MCP implementation used in this study is available at \url{https://github.com/zhou-zhichao/robotstudio-mcp}.} A RobotStudio add-in exposes a local HTTP API over port 8080. The MCP server translates tool calls from the language-model client into HTTP requests to the add-in, which then interacts with RobotStudio SDK objects and the virtual controller.
The bridge exposes 16 MCP tools covering station status, joint-state reading, simulation start/stop/reset, RAPID upload and execution, event-log retrieval, module/source inspection, RAPID variable access, I/O signal access, scene-object inspection, and screenshot capture.

\subsection{Validation cycle}
\label{subsec:validation}

The validation cycle starts after the generator produces a RAPID module. The workflow first applies structural checks to ensure that module and procedure delimiters are balanced and no template placeholders remain. A semantic validation step then checks whether user-specified objects, tools, I/O signals, and motion intent are represented. The MCP bridge subsequently uploads the code into RobotStudio, executes it on the virtual controller, and retrieves controller feedback.

When execution fails, the workflow uses the diagnostic output to revise the program. The main feedback signals used in this paper are controller execution status, event-log errors, joint readings, I/O signal values, and scene-object positions and bounding boxes. Scene introspection also helps avoid hardcoded assumptions. For example, the workflow can list available \texttt{wobjdata} declarations instead of manually deriving workobject coordinates, and it can read object bounding boxes to compute placement heights for different part sizes.

\section{Experimental Setup}
\label{sec:setup}

The evaluation combines retrieval benchmarking, code-generation checks, and RobotStudio simulation cases. Table~\ref{tab:evaluation-scope} states the unit of analysis for each part so that the results are interpreted at the correct scope. Retrieval metrics compare configurations over the same 30 queries; code-generation checks summarize observed module quality within logged generation sets; RobotStudio cases examine whether execution feedback can expose failures missed by text-level checks.

\begin{table}[htbp]
\centering
\caption{Evaluation scope and intended comparisons.}
\label{tab:evaluation-scope}
\begin{tabularx}{\linewidth}{p{0.24\linewidth}p{0.16\linewidth}Y}
\toprule
Evidence item & Unit & Role in the paper \\
\midrule
Retrieval benchmark & 30 queries & Direct comparison of retrieval configurations using the same query set. \\
Pure LLM baseline & 5 modules & Small diagnostic baseline for RAPID hallucination modes; not a paired statistical baseline for all generated modules. \\
RAG generation log & 21 modules & Broader check of structural validity and functional coverage after retrieval grounding. \\
Semantic review subset & 10 modules & Subset with logged full-pipeline review outcomes before simulation. \\
Release-height case & 1 cell task & Mechanism evidence that simulation detects a suction release failure missed by text checks. \\
Pyramid variants & 2 tasks & Case evidence for product-variant reconfiguration using the same RobotStudio cell. \\
\bottomrule
\end{tabularx}
\end{table}

The retrieval benchmark contains 30 queries constructed for pick-and-place operations, pallet flow, soldering sequences, signal processing, and error recovery logic. Each query was manually associated with relevant documentation and code chunks. In addition, 21 code-generation trials were used to evaluate RAPID module quality across seven categories, including path following, pick and place, gripper libraries, simple motion, arc welding, tool definition, and target-based programming.

The first RobotStudio case is a conditional pick-and-place task in which a suction tool handles boxes of different heights. This case was selected because the critical failure is not a syntax error: the generated program can compile and move to reachable targets while still failing to release the workpiece correctly in simulation.

The second RobotStudio case evaluates product-variant reconfiguration. Two 14-block pyramid tasks were run in the same RobotStudio cell with different product variants and target pallets. The green box variant was twice as tall as the orange box variant and was placed on a different pallet. This pair tests whether scene introspection and simulation feedback can support changed product geometry, placement heights, and robot configuration constraints.

Three system configurations were compared for the code generation part. The pure LLM baseline generated RAPID directly from the user query without retrieval. The naive RAG configuration used the dual retrieval streams but no query decomposition or semantic validation step. The full pipeline added query decomposition, structured prompt assembly, semantic review, and MCP-based simulation feedback. The metrics were selected to match the engineering risks in the workflow. Retrieval precision, recall, and mean reciprocal rank (MRR) indicate whether relevant manuals and templates enter the prompt context early enough to influence generation. For a query set $Q$, MRR was computed as:
\[
\mathrm{MRR}=\frac{1}{|Q|}\sum_{q \in Q}\frac{1}{\mathrm{rank}_q},
\]
where $\mathrm{rank}_q$ is the rank of the first relevant retrieved item for query $q$. Structural pass rate checks whether the output is a complete RAPID module, while content pass rate and validation issues check whether the program represents the requested tools, workobjects, I/O behavior, and motion intent. RobotStudio case studies were evaluated by generate-test-correct iterations, detected failure signal, correction source, and final execution outcome.

\section{Results}
\label{sec:results}

\subsection{Retrieval grounding and semantic validation}
\label{subsec:rag-results}

Table~\ref{tab:retrieval} shows the retrieval benchmark results. The document-only baseline achieved 0.52 precision and 0.58 MRR. Adding the code-template stream raised precision to 0.68 and MRR to 0.72. The improvement indicates that code examples help bridge the gap between user intent and formal RAPID terminology.

\begin{table}[htbp]
\centering
\caption{Retrieval performance averaged over 30 test queries.}
\label{tab:retrieval}
\begin{tabular}{lccc}
\toprule
Configuration & Precision & Recall & MRR \\
\midrule
Naive RAG (documentation only) & 0.52 & 0.41 & 0.58 \\
Dual retrieval with code templates & 0.68 & 0.57 & 0.72 \\
\bottomrule
\end{tabular}
\end{table}

The pure LLM baseline produced complete RAPID modules in all five baseline trials, but only one module passed content validation, giving a 20\% content pass rate within this diagnostic set. Typical failures included hallucinated instructions such as \texttt{DIRead} and \texttt{DInput}, misuse of \texttt{CTime()} as a numeric value, invalid signal declarations, redeclaration of built-in constants, and non-standard error handling syntax. These errors are plausible to a general coding model but invalid for ABB RAPID.

With retrieval grounding, all 21 generated modules passed structural validation. The full pipeline also improved functional coverage in the logged generation set: motion-instruction inclusion reached 100\%, and I/O instruction coverage increased from 36\% in naive RAG to 50\% in the full pipeline. Semantic review was logged for a subset of 10 full-pipeline modules; 4 were flagged for correction because of missing user-specified tools or workobjects, incorrect RAPID data type structures, non-compiling function misuse, missing section markers, unsafe motion patterns, or excessive speed and blending. These results are therefore reported as scoped quality checks rather than as one statistically balanced benchmark.

\subsection{Release-height validation in RobotStudio}
\label{subsec:pick-place-results}

The pick-and-place case demonstrates why static and semantic checks are insufficient for robot programs that interact with physical workpieces. The task used a suction tool to pick boxes from a conveyor and place them on a pallet. A release offset calibrated for a 100 mm orange box was reused for a 200 mm green box. The generated RAPID code remained structurally valid, and the motion targets were reachable, but the suction cup failed to release the green box because the tool-center point was geometrically embedded in the box during release.

\begin{figure}[htbp]
\centering
\includegraphics[width=0.7\linewidth]{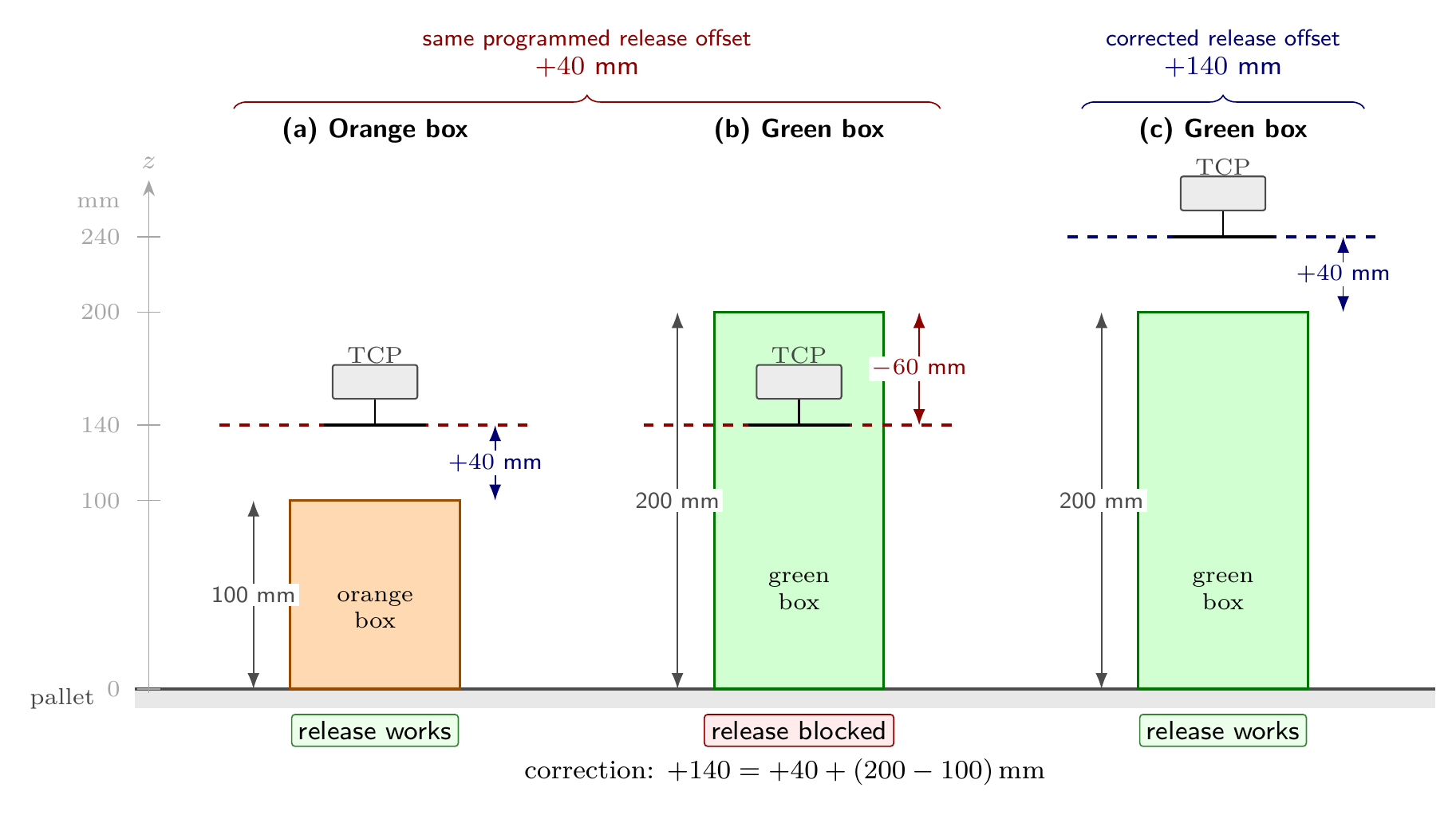}
\caption{Release-height failure exposed through RobotStudio simulation. The same +40 mm offset works for the 100 mm orange box but leaves the TCP inside the 200 mm green box. The corrected release offset is +140 mm.}
\label{fig:release-height}
\end{figure}

Figure~\ref{fig:release-height} shows the physical cause of the error. The programmed release plane was valid for the orange box but 60 mm below the top surface of the green box. The corrected release offset was computed as +140 mm, equal to the original +40 mm clearance plus the 100 mm height difference between the two product variants.

The failure signal was not a manual judgement from the screenshot. After the release command switched the suction output off, the workflow queried I/O state and scene objects through MCP. A placement was accepted only when the workpiece was no longer moving with the suction tool and its global scene position remained near the expected pallet pose. A failed release was recorded when the generated box did not detach or did not settle as an independent pallet object after the release command. Structural checks did not detect this failure because the module syntax and motion instructions were valid; semantic review only partially captured it because it did not model suction geometry. Screenshots were retained for documentation of the final state.

This case shows that the purpose of MCP is not only to upload code into RobotStudio, but to close a feedback loop around cell constraints. The diagnostic value comes from running the generated program in the same type of virtual controller and scene geometry that an engineer would use for offline validation.

\subsection{Product-variant reconfiguration}
\label{subsec:variant-results}

The product-variant reconfiguration case used two 14-block pyramid tasks in the same simulated manufacturing cell. In both variants, the workflow had to generate blocks from the conveyor, wait for sensor conditions, pick one block at a time, and place it into a 3-layer pyramid. The important difference is that the product and target pallet changed. The orange case used shorter boxes on the right pallet; the green case used taller boxes on the left pallet.

\begin{figure}[htbp]
\centering
\begin{tabular}{cc}
\includegraphics[width=0.47\linewidth]{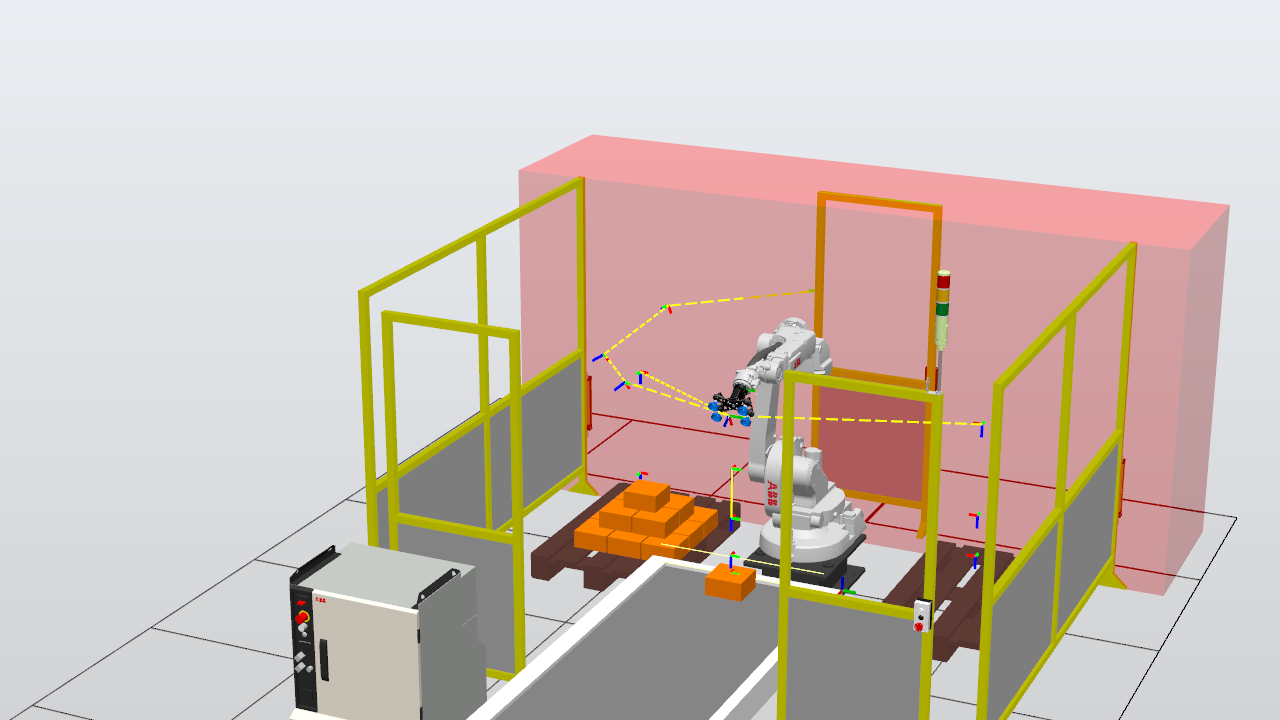} &
\includegraphics[width=0.47\linewidth]{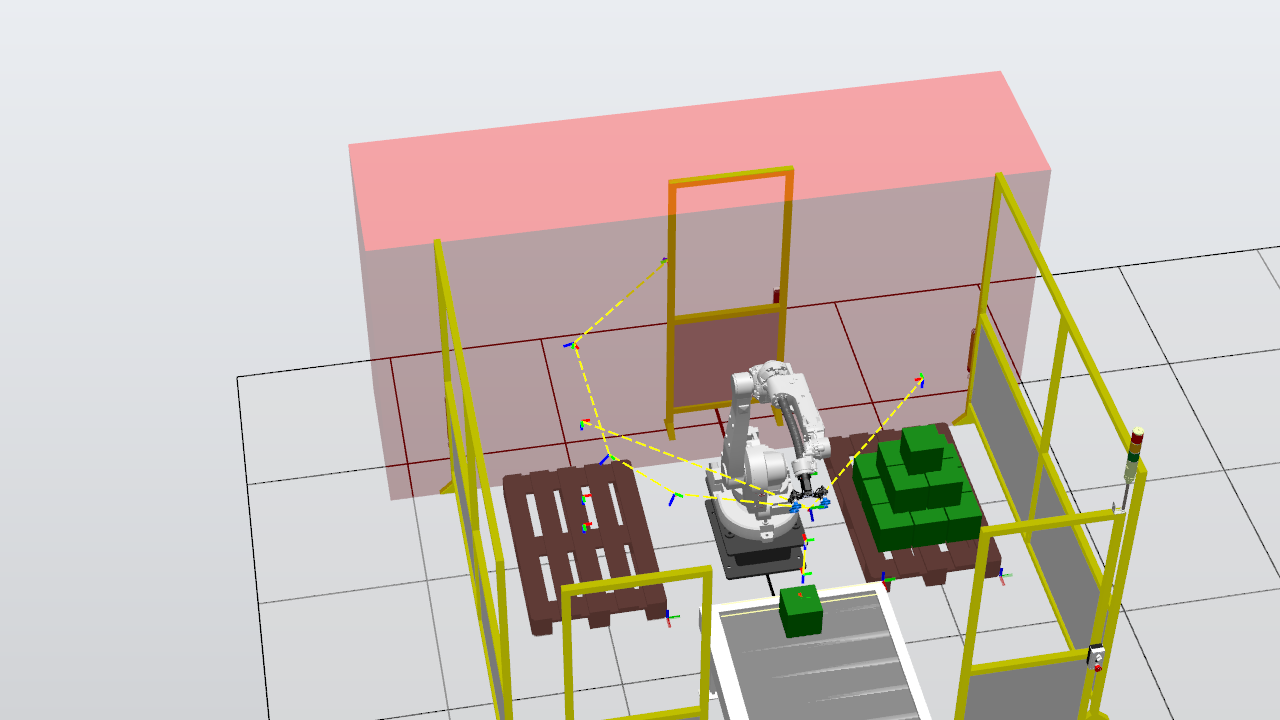} \\
(a) Orange-box variant, right pallet & (b) Green-box variant, left pallet
\end{tabular}
\caption{Product-variant reconfiguration in the RobotStudio cell. The orange variant succeeded on the first attempt after scene introspection; the green variant required additional recovery because the changed pallet and box geometry exposed reach and joint-configuration constraints.}
\label{fig:pyramid-results}
\end{figure}

Two MCP introspection tools were decisive in the orange-box variant. First, \texttt{list\_rapid\_variables} allowed the workflow to discover existing workobjects directly from the controller, avoiding manual coordinate reconstruction. Second, \texttt{get\_scene\_objects} returned object bounding boxes, allowing the workflow to read the orange box size as 200 mm $\times$ 200 mm $\times$ 100 mm and compute the placement heights for the pyramid layers. The orange 14-block pyramid succeeded on the first attempt.

The green-box variant changed both the box height and the placement side. The workflow again read the bounding box, now 200 mm $\times$ 200 mm $\times$ 200 mm, and recomputed the release and layer heights. The first two attempts failed because some placement targets were outside the feasible reach envelope. After the target base and approach height were adjusted, the third attempt placed six blocks but then triggered a controller-reported out-of-range error on the robot's fifth axis, one of the wrist-orientation axes. This was not a textual error in RAPID code; it emerged from the interaction between target placement, robot kinematics, and the joint configuration selected during repeated cycles.

\begin{figure}[htbp]
\centering
\begin{tabular}{cc}
\includegraphics[width=0.47\linewidth]{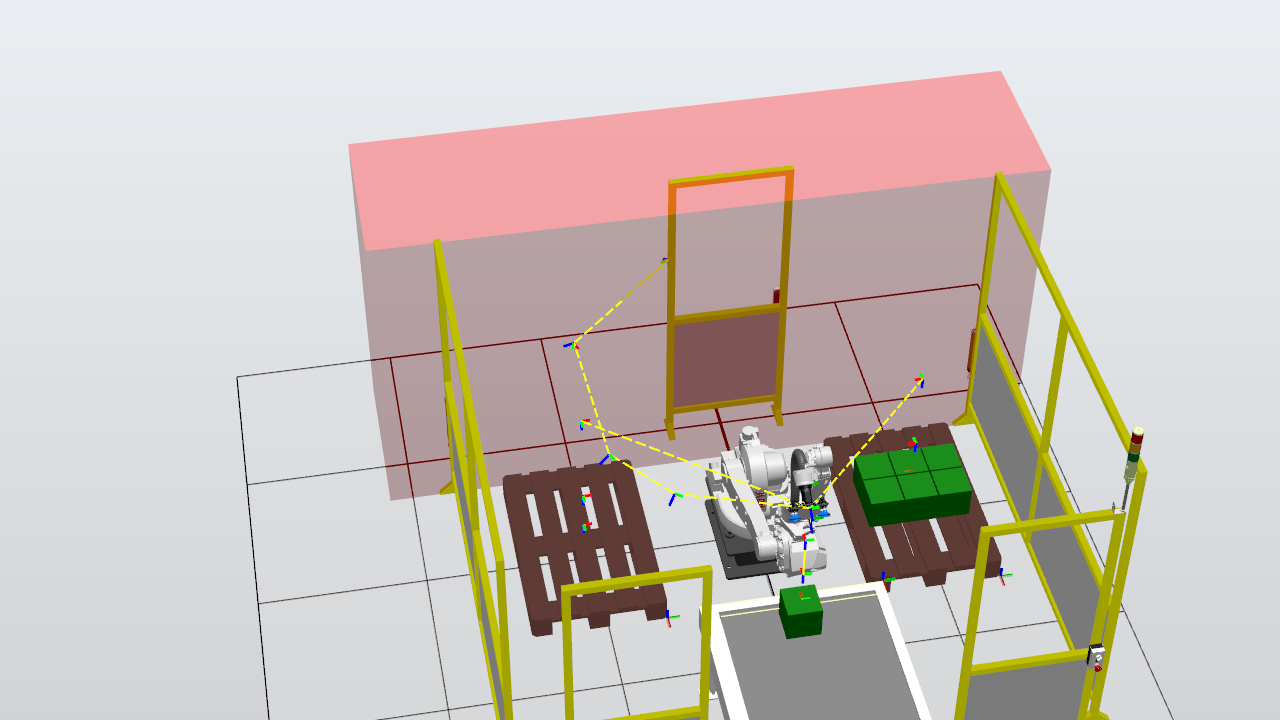} &
\includegraphics[width=0.47\linewidth]{fig_result_green_pyramid.png} \\
(a) Failure after wrist-axis drift & (b) Successful run after recovery pattern
\end{tabular}
\caption{Joint-configuration recovery in the green-box variant. The successful correction used \texttt{MoveAbsJ} to reset the robot to a calibration pose and \texttt{MoveJ} to return to the named home pose without forcing the fifth axis back near its stored high-angle home pose.}
\label{fig:wrist-recovery}
\end{figure}

The correction is shown in Figure~\ref{fig:wrist-recovery}. The unsuccessful versions returned to the stored home joint pose using \texttt{MoveAbsJ}; joint readings showed this pose placed the fifth axis at approximately 111.7 degrees. This value is reported as a high-angle recovery pose, not as the maximum angle reached during the failing path. The actual failure was identified from the RobotStudio controller error state and the experiment log entry reporting an out-of-range wrist-axis motion during the subsequent move. The successful version first used \texttt{MoveAbsJ} to move to a calibration pose with the fifth axis near zero, then used \texttt{MoveJ} to return to the named home pose so the controller could choose a safer joint solution. The final green-pyramid program placed all 14 blocks successfully.

\section{Discussion}
\label{sec:discussion}

The results suggest that industrial robot programming needs a different kind of LLM workflow from general robot-task planning. Code as Policies shows that LLMs can translate natural-language intent into executable control logic \cite{liang2023code}. ChatGPT-for-robotics shows a related potential for robot-oriented reasoning and programming, while also noting reliability limits \cite{vemprala2024chatgpt}. That framing is useful, but it often leaves the generated program separated from the vendor-specific controller, workobjects, I/O signals, and station geometry that determine whether the program will actually run. In contrast, the workflow in this paper treats RobotStudio feedback as part of the generation process rather than as a final manual check.

The RAG results are consistent with earlier work on RAG for robot code generation \cite{salunkhe2026rags}: grounding helps because RAPID programming contains many domain-specific names, data structures, and controller conventions that are unlikely to be reliably inferred from a general model alone. The present work extends that idea by adding an execution channel. The release-height and pyramid cases show that some errors only appear after the program is run in the configured RobotStudio station. This is an important distinction from retrieval-only approaches: RAG can reduce invalid code constructs, but it cannot by itself know whether a suction release height, a placement pose, or a repeated recovery motion is feasible in the active station.

The role of MCP in this workflow is to make RobotStudio feedback available through explicit interfaces instead of screen interaction or ad hoc scripts. The language-model client can request event logs, joint readings, I/O values, RAPID variables, and scene-object data as tool outputs, which makes the correction loop easier to inspect and reproduce. Compared with conventional automated testing for industrial robotic systems \cite{dossantos2024aat4irs}, the distinctive point is that the same feedback can be used not only to accept or reject a program, but also to guide the next code-generation attempt.

The human involvement boundary is also important. The reported workflow did not autonomously design the RobotStudio station, tools, workobjects, I/O mapping, or Smart Components; these were prepared before the experiments. The documentation set, templates, prompts, and task descriptions were also curated by the authors. During the reported validation loops, no manual line-by-line RAPID editing was counted inside the generate-test-correct cycle, but a human still initiated tasks, selected when to run candidate corrections, and confirmed final results using tool outputs and screenshots. The claim is therefore reduced manual program debugging, not fully autonomous cell engineering.

Several limitations remain. The evaluation is preliminary and uses a limited number of tasks and one RobotStudio cell. The pure LLM baseline contains only five trials, the semantic-review subset contains 10 modules, and no statistical significance test was performed. The RobotStudio evidence should therefore be read as case-based mechanism validation rather than a broad benchmark. Safety PLC integration, welding process constraints, and multi-robot coordination were not validated. The current bridge is also platform-specific: applying the method to KUKA, FANUC, or PLC environments would require corresponding MCP tool implementations. The system also does not yet persist learned corrections across sessions; product-specific release offsets, feasible placement envelopes, and recovery poses should eventually be stored in a persistent knowledge base and enforced as hard generation constraints.

\section{Conclusion}
\label{sec:conclusion}

This paper presented a workflow for generating and validating ABB RAPID programs by combining retrieval-grounded code generation with Model Context Protocol access to RobotStudio. The retrieval results show that combining technical documentation with production-style code templates improves access to relevant RAPID knowledge, while the generation checks show fewer domain-specific errors than an ungrounded LLM baseline. The RobotStudio cases demonstrate the added value of execution feedback: generated programs that are textually plausible can still fail because of release geometry, unreachable targets, or unsafe recovery motions. By exposing RobotStudio operations through MCP tools, the workflow connects natural-language task descriptions, grounded RAPID generation, simulation execution, and diagnostic correction in a single loop. The approach does not remove the need for a prepared station or expert supervision, but it provides a practical route toward faster reprogramming and debugging of industrial robot cells.

\section*{Acknowledgments}

The study was supported by Vinnova, Sweden's innovation agency, under grant number 2024-03234,
``Code Agents: AI-powered end-to-end solutions for flexible manufacturing.'' The work was carried
out within Chalmers' Area of Advance Production, whose support is gratefully acknowledged. The
computations were enabled by resources provided by the National Academic Infrastructure for
Supercomputing in Sweden (NAISS), partially funded by the Swedish Research Council through grant
agreement no. 2022-06725.

\bibliographystyle{elsarticle-num-names}
\bibliography{sample}

\end{document}